\documentclass[
]{ceurart}
\usepackage[utf8]{inputenc}
\usepackage{amsfonts}
\usepackage{amssymb}
\usepackage{amsmath}
\usepackage{amsthm}
\usepackage{xcolor}
\usepackage{hyperref}
\usepackage{algpseudocode}
\usepackage{algorithm}

\newcommand{\q}{\mathsf{q}}

\renewcommand{\emph}{\textbf}

\newcommand{\val}[1]{[\![{#1}]\!]}
\newcommand{\descr}[1]{(\![{#1}]\!)}
\renewcommand{\phi}{\varphi}
\begin{document}
\copyrightyear{2021}
\copyrightclause{Copyright for this paper by its authors.
  Use permitted under Creative Commons License Attribution 4.0
  International (CC BY 4.0).}

\conference{The 1st International Workshop on
Knowledge Representation for Hybrid Intelligence}

%\title{Interrogative agendas for classification and outlier detection with FCA}
\title{A Meta-Learning Algorithm for Interrogative Agendas}

\author[1,2]{Erman Acar}
\address[1]{LIACS, Universiteit Leiden}

\author[2]{Andrea {De Domenico}}
\author[2]{Krishna Manoorkar}
\author[2]{Mattia Panettiere}
\address[2]{Vrije Universiteit Amsterdam}
%\author{Erman Acar}

\begin{abstract}
Explainability is a key challenge and a major research theme in AI research for developing intelligent systems that are capable of working with humans more effectively. An obvious choice in developing explainable intelligent systems relies on employing  knowledge representation  formalisms which are inherently tailored towards expressing human knowledge e.g.,  \textit{interrogative agendas}. In the scope of this work, we focus on formal concept analysis (FCA), a standard knowledge representation formalism, to express interrogative agendas, and in particular to categorize objects w.r.t. a given set of features.  Several FCA-based algorithms have  already been in use for standard machine learning tasks such as classification and outlier detection. These algorithms use a single concept lattice for such a task, meaning that the set of features used for the categorization is fixed. Different sets of features may have different importance in that categorization, we call a set of features an agenda. 
%A framework for studying categorizations with different agendas using  FCA was developed in \cite{FLexiblecat2022}, however,
In many applications a correct or good agenda for categorization is not known beforehand. In this paper, we propose a meta-learning algorithm to construct a good interrogative agenda explaining the data. Such algorithm is meant to call existing FCA-based classification and outlier detection algorithms iteratively, to increase their accuracy and reduce their sample complexity. The proposed method assigns a measure of importance to different set of features used in the categorization, hence making the results more explainable. 
\end{abstract}


\maketitle

\section{Introduction}
As artificial intelligence (AI) technologies are playing key roles in our daily lives, developing intelligent systems which can work with humans more effectively (instead of replacing them) is becoming a central research theme \cite{9153877,peng2022investigations,russell2021human}.  Such theme is mostly pronounced as \textsl{hybrid intelligence}, aiming to benefit from the strengths of both humans and the machine intelligence in solving problems. Developing  systems of such capability demands fundamentally novel approaches to major research problems in AI: state-of-the-art systems outperform humans in many cognitive tasks from  playing video games \cite{hester_deep_2017} to pattern recognition \cite{liu2019comparison}, however they fall short when it comes to other tasks such as common sense reasoning, performing causal discovery, and behavioural human capabilities such as explaining its own decisions, adapting to different environments, collaborating with others, etc. A particular challenge in developing such systems relies on making them more interpretable \cite{9153877,tjoa2020survey,TIDDI2022103627} which is the main focus of this paper. 		

An obvious medium in making such systems interpretable relies on employing an existing knowledge representation  formalism which is inherently tailored towards expressing human knowledge. One such type of human knowledge that is relevant in problem solving is captured by the notion of \textit{interrogative agenda} (also called research agenda \cite{enqvist2012modelling}) of an epistemic agent (which will be explained further in detail in Section ~\ref{prelim: imterrogative agenda}). Intuitively, given a context, an interrogative agenda abstracts a set of features that an epistemic agent is interested in. In order to express interrogative agendas we employ the knowledge representation formalism of \textit{formal concept analysis}. 

Formal concept analysis (FCA) is an influential foundational theory in knowledge representation and reasoning \cite{priss2006formal, qadi2010formal, poelmans2010formal, valtchev2004formal, poelmans2013formal, ganter2012formal, wille1996formal} which provides a framework for categorizing objects w.r.t.~a given set of features.
The set of features used in the categorization (formal context in FCA) can be identified as its agenda, and different agendas will correspond to different categorizations. The agenda used to categorize a set of objects may be chosen on several factors like the availability and precision of the data, the categorization methodology, and the purpose of the categorization.\footnote{A logical framework for studying these different categorizations obtained from different agendas and their interaction was developed in our earlier work \cite{FLexiblecat2022} and applied to auditing domain.} In this paper, we focus on obtaining concept lattices (possibly fuzzy) corresponding to different agendas (possibly non-crisp)
%investigating their properties and some their possible employment in auditing.
However, in many  applications, it is unclear which interrogative agenda (Sec.~\ref{prelim: imterrogative agenda}) is best suited to obtain a categorization that can be useful in dealing with a given problem. Thus, in this work, we focus on the task of using a machine learning algorithm to learn such agendas, and hence a ``good categorization'' for the problem at hand. In particular, we will address the task of classification and outlier detection.

In the realm of machine learning, formal concept analysis has been used in the past for classification, outlier detection, rare concept mining and identification of rare patterns (Sec.~\ref{Sec:Classification and outlier detection using concept lattice}). However, to the best of our knowledge, all these methods use a single concept lattice  (or its sublattice) to deal with the problems mentioned above. That is, the agenda of the categorization is fixed beforehand. The main difficulty in using such techniques relies on the fact that there are exponentially many subsets of features (and weights) one has to take into account. 
%different importance of different sets of features. 
On the other hand, since some features may not be relevant for a given classification task, removing them can reduce the data collection cost, its complexity, and may even improve the accuracy for some tasks. However, determining the set of relevant features can be difficult, and it is an important part of the preprocessing phase for many such algorithms. 

In this paper, we propose a meta-learning algorithm to identify the best-suited agenda (and hence categorization). That is, to estimate the significance of different sets of features for the given task. The incorporation of such outer-loop on top of an existing classification or outlier detection algorithm can potentially increase its generalising power and the performance. Another major advantage of such method is that the learned agendas provide us an estimation of the importance of different sets of features for the given task, making our results more explainable. 

\paragraph{Structure of paper.} In Section  \ref{sec:preliminaries}, we provide the relevant preliminaries. In Section \ref{Sec:Classification and outlier detection using concept lattice}, we give an overview of FCA-based classification and outlier detection algorithms. In Section \ref{sec:Learning interrogative agendas}, we describe the framework for learning agendas and provide a generic learning algorithm. In Section \ref{sec:Conclusion}, we conclude and give some directions for future research. 
\section{Preliminaries}\label{sec:preliminaries}
\subsection{Formal concept analysis}
A {\em formal context} \cite{ganter2012formal}  is a structure $\mathbb{P} = (A, X, I)$ such that $A$ and $X$ are sets of {\em objects} and {\em features}, respectively, and $I\subseteq A\times X$ is the so-called {\em incidence relation} which records whether a given object has a given feature. That is, for any object $a$ and feature $x$, $a I x$ iff $a$ has feature $x$. 
Formal contexts can be thought of as abstract representations of e.g., databases, tabular data and such. %where elements of $A$ and $X$ represent objects and features, respectively, and the relation $I$ records whether a given object has a given feature. 
Every formal context as above induces maps $I^{(1)}: \mathcal{P}(A)\to \mathcal{P}(X)$ and $I^{(0)}: \mathcal{P}(X)\to \mathcal{P}(A)$, respectively defined by the assignments 
\begin{equation}
  I^{(1)}[B]: = 
\{x\in X\mid \forall a(a\in B\Rightarrow aIx)\},\quad 
 I^{(0)}[Y] = 
\{a\in A\mid \forall x(x\in Y\Rightarrow aIx)\}.
\end{equation}
A {\em formal concept} of $\mathbb{P}$ is a pair 
$c = (\val{c}, \descr{c})$ such that $\val{c}\subseteq A$, $\descr{c}\subseteq X$, and $I^{(1)}[\val{c}] = \descr{c}$ and $I^{(0)}[\descr{c}] = \val{c}$. 
A subset $B \subseteq A$ (resp.\ $Y\subseteq X$) is said to be {\em closed} or {\em Galois-stable} if $Cl(B)=I^{(0)}[I^{(1)}[B]]=B$ (resp.\ $Cl(Y)=I^{(1)}[I^{(0)}[Y]]=Y$).
The set of objects $\val{c}$ is intuitively understood as the {\em extension} of the concept $c$, while  the set of features $ \descr{c}$ is understood as its {\em intension}. 
The set   of the all formal concepts of $\mathbb{P}$ (denoted by $L(\mathbb{P})$) can be partially ordered as follows: for any $c, d\in L(\mathbb{P})$, 
\begin{equation}
c\leq d\quad \mbox{ iff }\quad \val{c}\subseteq \val{d} \quad \mbox{ iff }\quad \descr{d}\subseteq \descr{c}.
\end{equation}
%\redfootnote{Apostolos: Are $L(\mathbb{P})$ and $\mathbb{P}^+$ two notations for the same thing? Do we really need to introduce a notation for the set of concepts?}
With this order, $L(\mathbb{P})$ is a complete lattice, the {\em concept lattice} $\mathbb{P}^+$ of $\mathbb{P}$. 
\subsection{Interrogative agendas}\label{prelim: imterrogative agenda} 
%In epistemology and formal philosophy, an epistemic agent’s (or a group of epistemic agents’, e.g.~users’) interrogative agenda (or research agenda \cite{enqvist2012modelling}) indicates the set of questions they are interested in, or what they want to know relative to a certain circumstance (independently of whether they utter the questions explicitly). In general, in each context, interrogative agendas act as cognitive filters that block content which is considered irrelevant by the agent. Only the information the agent considers relevant is actually absorbed (or acted upon) by the agent and used e.g.~in the formation of their beliefs, etc.  Deliberation  and negotiation processes can be understood in terms of whether and how decision-makers/negotiators succeed in modifying their own interrogative agendas or those of other agents in the network, and the outcomes of these processes can be described in terms of the “common ground” agenda thus reached. Also, phenomena such as polarization \cite{myers1976group}, echo chambers \cite{sunstein2001republic} and self-fulfilling prophecies \cite{merton1948self} can be understood in terms of the formation and dynamics of interrogative agendas among networks of agents. 

In epistemology and formal philosophy, interrogative agenda (or research agenda \cite{enqvist2012modelling}) of an epistemic agent (or group of agents e.g.,~users) indicates the set of questions they are interested in, or what they want to know relative to a certain circumstance. 
%(independently of whether they utter the questions explicitly).
Intuitively, in any context, interrogative agendas act as cognitive filters that block content which is deemed irrelevant by the agent. Only the information the agent considers relevant is used
%actually absorbed (or acted upon) by the agent and used 
e.g.~in the formation of their beliefs, or actions, etc.  Deliberation  and negotiation processes can be described as whether or how agents succeed and interact in shaping their interrogative agendas, and the outcomes of these processes can be described in terms of the aggregated (or “common ground”) agenda.
Also, phenomena such as polarization \cite{myers1976group}, echo chambers \cite{sunstein2001republic} and self-fulfilling prophecies \cite{merton1948self} can be described in terms of the formation and dynamics of interrogative agendas among networks of agents.

%In classification tasks too, for a given set of objects,  we may have different agendas  serving different aims. For example, the agenda for classification of consumers for a grocery store based on their consumer preference is very different from the agenda of a political analyst trying to classify same set of people based on their political inclinations. Different agendas for different tasks and agents lead to different classifications of objects.
%which interest them. 

Dealing with a classification or outlier detection problem, we may have different agendas for different aims. For example, the agenda for the classification of consumers for a grocery store based on their buying preferences is very different from the agenda of a political analyst trying to classify the same set of people based on their political inclinations. Thus, interrogative agendas play an important role in determining natural or useful categorization for a specific purpose. 

\subsection{Interrogative agendas and flexible categorization}
\label{ssec:Interrogative agendas and flexible categorization}
Let $\mathbb{P}=(A,X,I)$ be a formal context. For a set of features $Y \subseteq X$, the formal context induced by $Y$ from $\mathbb{P}$ is $(A,X,I \cap A \times Y)$. Given the set of all the features $X$, the (non-crisp) interrogative agenda of an agent can be described by a mass function on $\mathcal{P}(X)$. For an agenda represented by $m:\mathcal{P}(X) \to [0,1]$, and any $Y \subseteq X$, $m(Y)$ represents the importance (or intensity of the preference) of the set of features $Y$ according to the agenda given by $m$. We assume that mass functions are normalized, that is, 
\begin{equation}
\sum_{Y \subseteq X} m(Y)=1.
\end{equation}
Any such mass function induces a probability or preference function $p_m: \mathcal{R} \to [0,1]$ such that $p_m((A,X,I \cap A \times Y))= m(Y)$, where  $\mathcal{R}$ is the set of all the formal contexts corresponding to the crisp agendas induced by subsets of $X$ (i.e. the formal contexts corresponding to each $Y \subseteq X$).

The agendas of different agents can be aggregated using different Dempster-Shafer rules \cite{shafer1992dempster, sentz2002combination, denoeux2006cautious} to obtain a categorization corresponding to aggregated agendas. A logical framework for deliberation between different agents having different agendas is developed in \cite{FLexiblecat2022}. This framework can be applied to study categorizations when different agents with different interests interact with each other for communication or joint decision making, as it is the case in auditing, community analysis, linguistics, etc. We also describe a method to approximate the importance of individual features from mass functions describing agendas by plausibility transform \cite{cobb2006plausibility} or pignistic transformation \cite{smets2005decision}, methods used in Dempster-Shafer theory to transform Dempster-Shafer mass functions to probability functions. These importance values of individual features can be useful in several  different applications like feature analysis, clustering, etc. 
\label{ssec:interrogativeag}

\section{Classification and outlier detection using concept lattices} \label{Sec:Classification and outlier detection using concept lattice}
In this section, we give an overview of different classification  
and outlier detection techniques using concept lattices.
\subsection{Classification using concept lattices}
Different algorithms have been applied to classify objects using formal concept analysis, that is, using concept lattices. 
 Fu et al. \cite{fu2004comparative} provide a comparison between  different FCA-based classification algorithms, such as  LEGAL \cite{liquiere1990legal}, GALOIS \cite{carpineto1993galois}, RULEARNER \cite{sahami1995learning}, CLNN and CLNB \cite{xie2002concept}. Prokasheva et al. \cite{prokasheva2013classification} describe different classification algorithms using FCA and challenges to such methods. 
 
 In \cite{kuznetsov2004machine}, Kuznetsov describes a classification algorithm that uses the JSM-method \cite{finn1989generalized,FINN1983351}. He proposes to use  concept lattices and training examples to form hypotheses as follows. Let $(A, X, I)$ be a formal context for the set of objects $A$ and the set of features $X$. We add an additional target feature $x \not\in X$ for denoting a class of an object. This partitions $A$ into three sets of objects $A_+$, $A_-$, and $A_\tau$ consisting of objects known to have feature $x$, objects known to not have feature $x$, and objects for which it is unknown whether or not they have it, respectively. Positive hypotheses for the JSM-method based on this formal context are given by the sets of features that are shared by a set of positive examples but not by any negative example. That is,  a set $H \subseteq X$ is a positive hypothesis iff $I^{(0)}[H] \cap A_+ \neq \emptyset$ and $H \not\subseteq  I^{(1)}[\{a\}] $ for any $a \in A_-$. Negative hypotheses are defined analogously. For any object $b$, it will be classified positively (resp. negatively) if $I^{(1)}[\{b\}]$ contains a positive (resp. negative) hypothesis but no negative (resp. positive) hypotheses. In case  $I^{(1)}[\{b\}]$ contains both or neither,  the classification is undetermined or some other method like majority voting can be used to classify $b$. The method sketched above has been used with different modifications in many FCA-based classification algorithms \cite{ganter2000formalizing,kuznetsov2013fitting,onishchenko2012classification}. Some classification algorithms based on FCA  use concept lattices to augment other classifiers like SVM \cite{carpineto2009concept},  Naive Bayes classifier and Nearest neighbour classifier \cite{xie2002concept} in preprocessing or feature selection. Other FCA-based classification methods include biclustering \cite{onishchenko2012classification}, and cover-based classification \cite{maddouri2004towards}. 
\subsection{Outlier detection using concept lattices}
Outlier detection can be considered as a special case of binary classification where the classes are outlier and non-outliers. Thus, any of the above-mentioned algorithms can be used for outlier detection using concept lattices. Some other methods or algorithms based on formal concept analysis have also been studied specifically for outlier detection or similar tasks like mining rare concepts or patterns  \cite{sugiyama2013semi,okubo2010algorithm,zhang2014outlier}. The simplest method to define the outlier degree of an element from a concept lattice is by using the size of its closure (i.e. the smallest category containing the element). Smaller size of closure of an object indicates that there are a small number of elements which have the same features as the object and thus it is likely to be an outlier. Sugiyama \cite{sugiyama2013semi} suggests  that the outlierness of an object in a concept lattice should not depend on the size of its closure but one must consider the number of concepts it creates. He suggests to define the outlierness score of a set of objects $B \subseteq A$ as
\begin{equation}
q(B): = |\{ (G,Y) \in \mathbb{P}^+ \mid B \subseteq G \, \text{or}\, I^{(1)}[B] \subseteq Y \}|.
\end{equation}
This definition is more suited to detect outliers that belong to a densely agglomerated cluster which locates sparsely if we overview the whole set of  objects. Zhang et al.~\cite{zhang2014outlier} propose an outlier mining algorithm based on  constrained concept lattices to  detect local outliers using a sparsity-based method. One of the key advantages of using formal concept analysis in classification or outlier detection over other algorithms is that FCA can be used to deal with both continuous and discrete attributes simultaneously, through the discretization of continuous attributes by conceptual scaling (Sec. \ref{ssec:COnceptual scaling}).

One of the major issues in applications of formal concept analysis is the complexity of the algorithms involved. The fundamental reason behind the high complexity is that in the worst-case scenario the number of categories in a concept lattice grows exponentially with the number of objects and features involved. Several techniques have been devised in past to overcome this complexity problem \cite{cole1999scalability,dias2010reducing,singh2017concepts}. 

\subsection{Discretization of continuous attributes and conceptual scaling} \label{ssec:COnceptual scaling}
In order to apply formal concept analysis on attributes with continuous values, we  need to discretize them.  The process of converting many-valued (possibly continuous-valued) attributes into binary attributes or features for FCA is known as conceptual scaling \cite{ganter1989conceptual}. Scaling is an important part of most FCA-based techniques and has been studied extensively \cite{ganter1989conceptual,prediger1997logical,prediger1999lattice}. Choosing the correct scaling method depends on the specific task the concept lattice is used for. 
\section{Learning interrogative agendas} \label{sec:Learning interrogative agendas}
Formal concept analysis categorizes a given set of objects w.r.t~a given set of features. Thus, the outlier detection (or the classification) task at hand depends on the features (or attributes) under consideration. However, in many applications it is hard to estimate which features are of importance and how important they are, that is the correct agenda, for a given task. Here we describe a machine learning framework that tries to solve this problem by using machine learning to learn a ``good'' agenda for the given task. This provides a way to improve the performance of FCA-based classification or outlier detection algorithms  by choosing the correct agenda. This also makes results more explainable by providing the importance value of each set of features. 
\subsection{Space of possible agendas}
As discussed in Section \ref{ssec:Interrogative agendas and flexible categorization}, an (non-crisp) interrogative agenda on a given set of features $X$ is given by a mass function $m:\mathcal{P}(X) \to [0,1]$, where for any $Y \subseteq X$, $m(Y)$ denotes the importance of the set of features $Y$ in the categorization.
%defined by the agenda given by $m$. 
The mass function $m$ induces a probability function $p_m:\mathcal{R} \to [0,1]$, where $\mathcal{R}$ is the set of all the (crisp) formal contexts induced from $\mathbb{P}=(A,X,I)$ by different crisp agendas i.e. subsets of $X$. For any categorization (formal context) $\mathbb{P} \in \mathcal{R}$, $p_m(\mathbb{P})$ denotes the likelihood assigned or preference given to the categorization  $\mathbb{P}$ by the agenda $m$. Thus, the set of all  possible non-crisp categorizations  (resp. non-crisp agendas)  induced from a context $\mathbb{P}$ is given by the set of all the probability functions on $\mathcal{R}$ (resp. the set of all the possible mass functions on $\mathcal{P}(X)$). As discussed in the introduction,  we want to learn  a ``good'' agenda that
%provides a ``good'' agenda 
leads to a categorization that can be used to complete a given task effectively. This corresponds to learning  a probability function $p$ on $\mathcal{R}$ which represents a suitable categorization for the given task. That is, we use machine learning to search for a ``good'' function in the space of probability functions on $\mathcal{R}$.
%For computational and mathematical simplification, here we propose the following simplifications.
For the sake of computational and notational convenience, here we propose the following simplifications.

Let  $\mathbb{R}$ be the set of real numbers. Let $f:\mathcal{R} \to \mathbb{R}$ be a map assigning weight $w_\mathbb{L} \in \mathbb{R}$ for every $\mathbb{P} \in \mathcal{R}$. For any $\mathbb{P} \in \mathcal{R}$, $f(\mathbb{P})$ denotes the importance (or preference) assigned to the context $\mathbb{P}$ or  to the corresponding set of features $Y$, where $\mathbb{P}=(A,X, I \cap A \times Y)$. We call any such function $f$ a non-crisp agenda as it gives weights (representing importance) to different sets of features. Any such function can be seen as a real-valued vector of dimension $|\mathcal{R}|$. Thus, the set of all such functions is isomorphic to the space $\mathbb{R}^{|\mathcal{R}|}$.  As this space is linear, the shift from  probability functions on $\mathcal{R}$ to real-valued functions simplifies the task of learning an agenda  (weight function) that minimizes loss using a simple gradient descent method. 

The weights  assigned to lattices can be interpreted as probabilities on $\mathcal{R}$, (and hence  mass functions on $\mathcal{P}(X)$) via normalization when all the weights are non-negative. The negative weights suggest that the corresponding   categorization is opposite to the preferred categorization for the task at hand. For example, suppose we are interested in detecting elements  with a value  of feature $f_1$ being abnormally  high, while the outlier detection method used finds 
outliers with value  of $f_1$ low. Then the learning algorithm is likely to assign a negative weight to the  agenda $\{f_1\}$. 

As discussed earlier, one of the major problems in applications of formal concept analysis is the complexity of the algorithms involved.  Here, we are proposing to consider priority (or weight) functions on a set of different concept lattices corresponding to different agendas. As the number of different (crisp) agendas induced from a set $X$ of features is exponential in $|X|$, this may add another exponential factor to the complexity of the algorithm. In many applications where the number of features is large, this may make the problem computationally infeasible. Thus, in most applications we  need to choose a smaller set of concept lattices or (crisp) agendas as a basis, that is set of (crisp) concept lattices on which the weight functions  are defined. We propose the following strategies for this choice. 

\begin{enumerate}
    \item \textbf{Choosing agendas that consist of a small number of features} In this strategy, we choose the (crisp) agendas consisting of  $\alpha$ or a smaller number of features to construct basis concept lattices for some fixed  $\alpha\ll |X|$. This is based on the idea that tasks like classification or outlier detection can be performed with good accuracy by considering only a small number of features together. This is especially the case with tasks involving epistemic components as humans use a limited number of features in combination for basic tasks like comparison and classification. As these agendas  consist of a small number of features, the number of concepts in these concept lattices is small. This makes the computational complexity low for most algorithms operating on concept lattices. Thus, this method can be applied for finding agendas when the algorithms may have high computational complexity for lattices with a large number of concepts.  In some situations, it may also be useful to add the full concept lattice (lattice corresponding to the full feature set $|X|$) to the set of basis lattices. This allows us to consider the full concept lattice with all available information for the task at hand while having the possibility of giving higher or lower (compared to other features) importance to some small subsets of features. For example, if the weights attached to all the lattices except those given by agendas $\{f_1\}$ and $X$ are close to $0$ and the weights assigned to these agendas are similar, it corresponds to the agenda in which the set of all features and $\{f_1\}$ are the only important sets of features. Thus, the concept lattice based on $f_1$ alone would be of high significance.  
    
    \item  \textbf{Choosing important  agendas based on prior or expert knowledge} For some tasks, we may have prior or expert knowledge  assigning different importance or priority to some lattices or agendas. In such cases, these lattices are taken as the set of basis lattices.  This provides us a way to incorporate prior or expert knowledge with other algorithms using formal concept analysis. 
    \item  \textbf{Choosing agendas adaptively} In this strategy, we start with a set of agendas given by all the sets  consisting of less than $\alpha$ features for some small $\alpha$ (usually taken as 1). We use machine learning to learn weights assigned to them, and then drop all the oness which get assigned a very low weight (after normalization).  We then consider agendas consisting of any  set of features that is a subset of the union of agendas that are not removed in the first step. Choosing  these agendas can be interpreted as considering combinations of features that are deemed important in the first learning step.  We then repeat the learning process with this new set of lattices. We keep repeating this process until all the agendas (lattices) added in the last step get assigned low weights or we reach $|X|$ (full concept lattice). In this way, we recursively check the possible combinations of  agendas  deemed to be important so far in the next recursive step. This method works on assumption that if a feature is not important on its own, then it is unlikely to be part of a set of features that is important. However, this assumption may fail in several situations. In such cases, this method should not be used to choose a basis. 
    \end{enumerate}
    There can be other effective strategies for choosing basis lattices for different tasks and algorithms. 
    \subsection{Learning algorithm}
    Once the set  of possible agendas (or concept lattices) is chosen, we apply some 
    classification or outlier detection algorithm on each of these. For every lattice $\mathbb{L} \in \mathcal{R}$, we start assigning it a random weight $w \in \mathbb{R}$. Let $Alg$ be any algorithm which performs classification or outlier detection for a fixed concept lattice.
    
    Suppose $Alg$ is a classification (resp. outlier detection) algorithm classifying a set $A$ of objects into $n$ classes using concept lattices. For any object $a$ and a class $k$, let $Alg_k(a, \mathbb{L})$ (resp. $Alg(a, \mathbb{L})$ denote the membership of  the object $a$ into the class $k$ (resp. outlier degree) according to the classification algorithm $Alg$ acting on the lattice $ \mathbb{L}$. Notice that we allow for our classifiers (resp. outlier detection algorithms)  to be interpreted as fuzzy or probabilistic such that membership value (resp. outlier degree) of $a$ belongs to $[0,1]$. For an algorithm $Alg$ with crisp output, the value  $Alg_k(a, \mathbb{L})$ (resp. $Alg(a, \mathbb{L})$) will be either $0$ or $1$.  For a given weight function $w: \mathcal{R} \to \mathbb{R}$, we say that the membership of $a$ in the class $k$ (resp. outlier degree of $a$) assigned by the algorithm $Alg$ acting on a non-crisp categorization described by $w$ is 
    \begin{equation}
    \label{eqn:outputs}
    out_k(a,w) = \frac{\sum_{\mathbb{L} \in \mathcal{R} } w(\mathbb{L})Alg_k(a, \mathbb{L})}{\sum_{\mathbb{L} \in \mathcal{R} }w(\mathbb{L})}.
    \end{equation}
    Intuitively, this corresponds to taking the weighted sum of the result given by $Alg$ on lattices with weights provided by the agenda $w$. Let $loss$ be a loss function for a given classification task, and let $loss(out)$   be the total loss for the classification (resp. outlier detection) when classes (outlier degrees) are assigned by $ Alg_k(a,w) $ (resp. $ Alg(a,w) $). We use a gradient descent method to learn the agenda $f_0$ that minimizes the loss. We then use the learnt agenda $f_0$ to assign a class to an object that is for any test object $b$, its predicted membership in class $k$ (resp. outlier degree) is $ Alg_k(b,f_0) $ (resp. $Alg(b, f_0)$). 
    
\begin{algorithm}
\footnotesize
\caption{Meta-Learning Algorithm for Interrogative Agendas}
\hspace*{\algorithmicindent} \textbf{Input:} a set of objects $A$, a set of features $X$, a training set $T\subseteq X$, and a map $y:T \to C$ representing the labels on the training set, an algorithm $Alg$ that takes in input some object and a concept lattice in $\mathcal{R}$, and outputs an element in $\mathbb{R}^C$ representing its prediction for each class; a loss function $loss$ that compares two classifications and outputs a real number, and a number of training epochs $M$.\\
\hspace*{\algorithmicindent} \textbf{Output} A model that classifies objects in $X$.
\begin{algorithmic}[1]
\Procedure{Train}{$A$, $X$, $T$, $y$, $Alg$, $loss$,  $M$}
    \State $\mathbb{L}_1,\ldots,\mathbb{L}_n \leftarrow $ \textbf{compute} the concept lattices of $\mathcal{R}$
    \State \textbf{let} $predictions$ be an empty map from $A$ to $\mathbb{R}^C$
    \State \textbf{let} $w$ be an array of weights of length $n$ initialized with random values in $\mathbb{R}$
    \For{$e = 1, \ldots, M$ } 
        \For{$a \in X$, $k \in C$}
            \State $predictions[a][k] \leftarrow \frac{\sum_{i = 1}^n w(\mathbb{L}_i)Alg_k(a, \mathbb{L}_i)}{\sum_{i = 1}^n w(\mathbb{L}_i)}$
        \EndFor
        \State \textbf{update} $w$ with an iteration of gradient descent using $loss(predictions)$
    \EndFor
\EndProcedure
\end{algorithmic}
\end{algorithm}
A generic  algorithm for outlier detection can be given in a similar manner. 

\subsection{Example} 
Let us consider the following toy data table providing some information on different types of apples. It contains information on the color, size, sweetness, and origin of the apples. We assume that all apples under consideration are either green or red. For conceptual scaling,  we divide sweetness, price, and volume into low, medium, and high. This converts these continuous-valued attributes into discrete-valued. The set of features is obtained by considering each value of attributes as a different feature. For example, High volume, red color, Medium price are a few of them. 
%\mpmnote{Why do we specify {\bf Volume}? I mean, Low, Medium, and High are not in $cm^3$. You are right of course. }

\begin{table*}[h]
    \label{Table:data table}
    \centering
    \begin{tabular}{c c c c c c}
       \hline
       \textbf{Type}& \textbf{Color}&\textbf{ Volume}&\textbf{ Sweetness} &\textbf{ Local}& \textbf{Price}\\
       \hline
       1 & red & High  & High & Yes & Medium\\
       2 & green & High & High & Yes & Medium \\
       3 & red & Medium & Medium & Yes& Medium\\
       4 & green & Low & High & No& Medium\\
       5 & green & High & Medium & No & Low\\
       6 & red & Medium & Low & Yes& Low \\
       7 & green & High & Medium & Yes &Low \\
       8 & green  & High & Medium & Yes& High\\
       \hline
    \end{tabular}
     \caption{Exemplary data table containing the information on different types of apples w.r.t. attributes such as color, volume, sweetness, et cetera. }
\end{table*}

Let $A$ and $X$ be the set of all types of apples and features respectively. The (non-crisp) agendas of interest to us are the ones assigning mass to an attribute and not to an individual feature. That is, we consider basis lattices corresponding to feature sets that contain all the values for a given many-valued attribute. As an example, if a $Y \subseteq X$ in the agendas corresponding to a  basis lattice   contains the feature high volume, then it must also contain the features low and medium volume. We use volume to denote the set of features \{high volume, low volume, medium volume\}. A similar convention is used for the other attributes as well.

Let  $I \subseteq A \times X$ be the incidence relation and
let $P$ be a customer. Suppose we are interested in classifying apples into types customer $P$ likes (class 1) and  does not like (class 2).  Given a formal context (concept lattice) $\mathbb{P}= (A, X, I \cap A \times Y)$, describing a categorization of these types of apples for a given agenda of interest $Y$,  we use the following simple algorithm to predict the class for a new type of apple. 
Let $A_+$  and $A_-$ be the set  of apples known to be in class 1 and  class 2  respectively (from the training set). A set of features $H \subseteq Y$ is said to be a positive (resp. negative) hypothesis in w.r.t.~a lattice $(A, X, I \cap A \times Y)$ iff $H$ is Galois-stable,  $I^{(0)}[H]$ is non-empty,  and $I^{(0)}[H] \cap A_-=\emptyset$ (resp.~$I^{(0)}[H] \cap A_+=\emptyset$). For any new element $t$, 
we put it in class 1 (resp. class 2) if the category $ I^{(1)}[\{t\}]$  contains only positive (resp. negative) hypothesis. The algorithm is inconclusive when it contains neither type of  hypothesis (no information) or contains both type of hypotheses (inconsistent information). 

 Suppose the classification of apples of types 1-8 into classes 1 and 2 for customer $P$ are as $Class \, 1=\{1,2,4,5, 7\}$ and $Class\,  2=\{3,6,8\}$ and suppose also that we use the full concept lattice (that is,  agenda $Y=X$). Let $t_0$ be a new type of apple that is green, has high volume, high sweetness, is local, and has a high  price. Consider   hypotheses $H_1=$ \{High sweetness\} and  $H_2=$ \{Green, local\} which are both contained in $ I^{(1)}[\{t_0\}]$. The hypothesis $H_1$ is positive while $H_2$ is negative. Thus, the above classification algorithm can not classify this object as the available information is inconsistent.  However, in many cases, some subsets of features are of much more importance to a  customer than others. For example, from the above classification, it is hinted that the customer $P$   considers Sweetness and Price as more important features than color or location. Our algorithm for learning agenda can help to  recognize this  difference in importance of different sets of features and allow us to classify such elements.
 
 Suppose we use our method to find the best categorization (or agenda) for the completion of this task using the above classification algorithm with the basis lattice consisting of lattices given by agendas comprising of one attribute (as discussed earlier, one attribute can correspond to multiple features due to scaling). We start with  random weights assigned to each of these lattices. We then use the classification algorithm described above to classify new types of apples into classes 1 and 2 using each of these lattices. We then sum over the weights of lattices in which elements are assigned to either class. The new object is assigned to the class which has a higher mass (the algorithm is indecisive if such a class does not exist). We use machine learning (gradient descent) to train the algorithm to find the best weights for this classification task. 
 
During the training phase, our algorithm can (generally) learn that the attribute  (or set of features) \{sweetness\} matters much more than other features to the customer $P$. Thus,  a high weight will be attached to the lattice with agenda \{sweetness\}. Thus, the above algorithm in combination with our method assigns  $t_0$ to class 1. 
Adding this method on top of a classification algorithm may give a  better classification (that is, more elements classified correctly with a given amount of training samples) given  our learnt information  `sweetness is much more important for $P$ in decision-making' is true. 

Similarly, higher (resp. lower) masses attached to   agendas consisting of different sets of a single attribute are helpful in better categorization when this attribute is  more (resp. less) important for the customer. Thus, using machine learning techniques  (for example, gradient descent when possible)  to learn the best possible agenda to provide categorization to complement the classification algorithm can improve its accuracy with less training. Considering more basis lattices may further improve the accuracy and sample complexity. For example, it can be seen that the types 5 and 7, which have medium sweetness and low price, belong to class 1. This provides us with another likely hypothesis that the customer likes apples that are of medium size (not necessarily high) but have a low price. This hints to us  that the agenda \{size,  price\} may be of significant importance to the customer. In case this agenda is indeed more important to the agent, the learning   would  assign it a high weight during training and  thus allows us to make  more accurate predictions with a fewer number of samples. However, an increasing number of basis lattices may increase computational complexity significantly, meaning that such a decision needs to be made judiciously. 

This simple example shows that the classification  algorithm described above can be improved in terms of accuracy, sample complexity, and explainability by adding a learning step for finding out the best agenda for categorization. Adding this step to the  different algorithms discussed in Section \ref{Sec:Classification and outlier detection using concept lattice}, used for classification and outlier detection using concept lattices can improve these algorithms in a similar manner.  This is especially the case for the tasks in which the importance of different features may be hard to estimate beforehand. The obtained  agendas can be defined formally using the   logical framework described in \cite{FLexiblecat2022}. In that paper, a logical model was used to represent deliberation between agents with different agendas. This framework can also be used to model deliberation or interaction between different learning algorithms by aggregating learnt agendas using different techniques described in \cite{FLexiblecat2022}. The agendas inferred by our learning  algorithm can be used for  further tasks  like aggregation from different sources. For example, if for two different classifiers the agendas learned are given by the mass functions $m_1$ and $m_2$ on $\mathcal{P}(X)$, then a combined classifier that takes both into account can be obtained by choosing the agenda $F(m_1, m_2)$, where $F$ is a suitable Dempster-Shafer combination rule \cite{sentz2002combination,smets1993belief,denoeux2006cautious}, and then applying  the classification algorithm to the resulting lattice. 

\section{Conclusion and future directions} \label{sec:Conclusion}
%In this paper, we continue our research on categorizations given by different features of importance, that is different agendas,  in the framework of formal concept analysis. 
In this paper, targeting the explainability line of hybrid intelligence research~\cite{9153877}, we proposed a meta-learning algorithm to learn a "good" (interrogative) agenda for categorization  (which is used by a potential FCA-based classification or outlier detection algorithm).  Adding such a learning step to a given algorithm  allows us to improve the accuracy and sample complexity of the procedure while also making it more explainable. On the empirical side, a performance evaluation and the ablation study on the results of employing different FCA-based classification and outlier detection algorithms is an avenue of future research. Another investigation line is the transferability analysis of  "good" agendas e.g., how much knowledge do we transfer and how good the data efficiency is when such an agenda is used on previously unseen environments/categorizations.  Noteworthy is extending this methodology towards other interesting application domains such as knowledge discovery, data visualization, information retrieval, etc.

%The interaction between the learnt agendas by the method discussed in this paper  and corresponding categorizations can be studied formally using the  framework developed in \cite{FLexiblecat2022}. 
On the theoretical side, this framework can be used to  model deliberation between agendas learnt from different algorithms, providing us a way to study their interaction, comparison, or combination. Within the interpretation of taking the concept lattice as expert knowledge,  the learnt agendas can also be aggregated or compared with agendas of different experts allowing us to incorporate learning and expert knowledge in categorization. From a multiagent systems perspective, it is especially useful to model subjective categorizations involving multiple agents (human experts and algorithms) with different agendas or goals interacting with each other.  In future work, we are considering investigation in a variety of directions e.g., investigating desirable properties of various aggregation mechanisms, representational power such as proportionality and fairness of induced agendas for multiple parties,  convergence and robustness guarantees for the "good" agendas, computational complexity analysis on hard and easy cases for (non-)crisp agendas, and extending our method on a more general framework in order to tackle the problem of features selection in an uniform way.

The meta-algorithm described in the present paper is currently employed in the development of an outlier detection algorithm with good results. Currently, it has been tested on the datasets from the ELKI toolkit \cite{Campos2016} and it has been compared against the algorithms discussed in it. A detailed report of the results will be available in the future.
%In future work, we want to apply this joint framework to develop a model for categorization  based on different agendas learnt from machine learning algorithms, and provided by different experts. These categorizations can be  used  for  different applications like outlier or anomaly detection, knowledge discovery, and community analysis. 


\begin{acknowledgments}
 Erman Acar is generously supported by the Hybrid Intelligence Project which is financed by the Dutch Ministry of Education, Culture and Science with project number 024.004.022. Krishna Manoorkar is supported by the NWO grant KIVI.2019.001 awarded to Alessandra Palmigiano.
\end{acknowledgments}

\bibliography{ref}
%%
%% Define the bibliography file to be used
\appendix
\end{document}